\documentclass[10pt,twocolumn,letterpaper]{article}

\usepackage{cvpr}
\usepackage{times}
\usepackage{epsfig}
\usepackage{graphicx}
\usepackage{amsmath}
\usepackage{amssymb}

\usepackage[pagebackref=true,breaklinks=true,letterpaper=true,colorlinks,bookmarks=false]{hyperref}

\cvprfinalcopy % *** Uncomment this line for the final submission

\def\cvprPaperID{****} % *** Enter the CVPR Paper ID here

\begin{document}

%%%%%%%%% TITLE
\title{Domain Adaptation for Real-World Single View 3D Reconstruction}

\author{Brandon Leung\\
UC San Diego\\
{\tt\small b7leung@ucsd.edu}
\and
Siddharth Singh\\
UC San Diego\\
{\tt\small sisingh@eng.ucsd.edu}
\and
Arik Horodniceanu\\
UC San Diego\\
{\tt\small ahorodni@eng.ucsd.edu}
}

\maketitle
%\thispagestyle{empty}

%%%%%%%%% ABSTRACT
\begin{abstract}
Deep learning-based object reconstruction algorithms have shown remarkable improvements over classical methods. However, supervised learning based methods perform poorly when the training data and the test data have different distributions. Indeed, most current works perform satisfactorily on the synthetic ShapeNet dataset, but dramatically fail in when presented with real world images. To address this issue, unsupervised domain adaptation can be used transfer knowledge from the labeled synthetic source domain and learn a classifier for the unlabeled real target domain.To tackle this challenge of single view 3D reconstruction in the real domain, we experiment with a variety of domain adaptation techniques inspired by the maximum mean discrepancy (MMD) loss, Deep CORAL, and the domain adversarial neural network (DANN). From these findings, we additionally propose a novel architecture which takes advantage of the fact that in this setting, target domain data is unsupervised with regards to the 3D model but supervised for class labels. We base our framework off a recent network called pix2vox. Results are performed with ShapeNet as the source domain and domains within the Object Dataset Domain Suite (ODDS) dataset as the target, which is a real world multiview, multidomain image dataset. The domains in ODDS vary in difficulty, allowing us to assess notions of domain gap size. Our results are the first in the multiview reconstruction literature using this dataset.
\end{abstract}

\begin{figure*}
\begin{center}
\includegraphics[width=14cm]{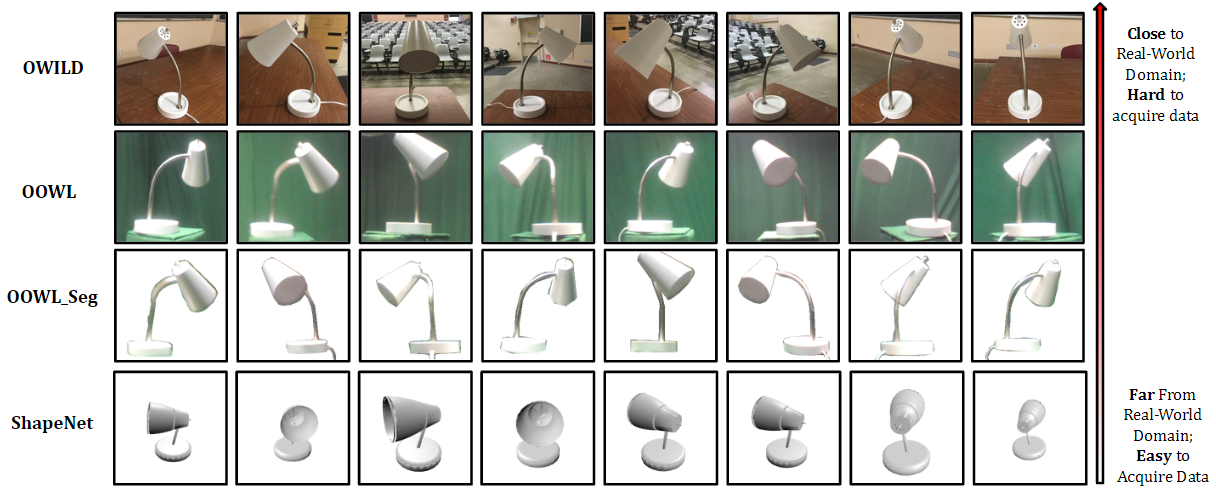} \\
\end{center}
   \caption{An example object (lamp) from datasets used for our project. OWILD, OOWL, and OOWLSeg are part of the ODDS dataset, and are real. Meanwhile, ShapeNet is synthetic.}
\label{fig:datasets}
\end{figure*}

%%%%%%%%% BODY TEXT
\section{Introduction} \label{sec:intro}

Humans are able to understand the visual world in 3D. This helps us manipulate objects and intuitively make sense of the world. Crucially, we don't need to observe everyday objects from all viewpoints in order to have this ability. Usually only one views will suffice for us to have a strong understanding of the 3D object -- for example, we can imagine the object from different novel viewpoints and ``mentally rotate'' them in our heads. In computer vision, this task is called \textit{single view 3D reconstruction}: given a single 2D input image, we wish to output 3D object models (eg in the form of voxels, point clouds, or triangular meshes). Note that this is different from the classical problem of structure from motion, which requires a dense, full coverage of all viewpoints ~\cite{ozyecsil2017survey}. The single view case is difficult because in general, it is highly unconstrained. For instance, given the view of picture of a car from the front, no algorithm or person can produce an image of the car from the back with 100 percent certainty. However, there are certain clues which can lead to a reasonable result since we know that the world follows certain geometric patterns and rules. Additionally, since we have seen many pictures of cars in the past, we can incorporate our prior knowledge specific to the car class.

Regardless, nearly all of them use a Synthetic dataset of mesh models called ShapeNet ~\cite{chang2015shapenet} for both training and testing. This is because it provides many 2D rendered images of objects, as well as their ground truth 3D representations. However, because they are all CAD models with little detail, no texture, and no background, models trained on ShapeNet do not perform well when presented images in the real world due to a domain gap between synthetic and real. To be practical, this model should be able to work with images in the real world, with complicated backgrounds. Therefore, techniques from the domain adaptation literature can be applied. This would allow us to transfer knowledge from a synthetic source domain with 3D ground truth to an real target domain without 3D ground truth.

In this paper, we present results which extend the work of a current state-of-the-art, synthetic-based, single view voxel reconstruction method called pix2vox ~\cite{xie2019pix2vox}, so that it can be applied in the real world. First, we provide a summary of the related literature in Section \ref{sec:related}. Then, our rationale for choosing this framework to work on is discussed in Section ~\ref{sec:methods}. In particular, we utilize several domain adaptation methods based on the maximum mean discrepancy (MMD) loss, Deep CORAL, and the domain adversarial neural network (DANN). We also propose a novel architecture which takes advantage of the fact that in this setting, target domain data is unsupervised with regards to the 3D model but supervised for class labels. Then, we share our results in Section ~\ref{sec:experiments}. We demonstrate that as is, pix2vox fails with real-world images, evaluate the domain adaptation methods, and verify the usefulness incorporating class labels.

\begin{figure*}
\begin{center}
\includegraphics[width=\linewidth]{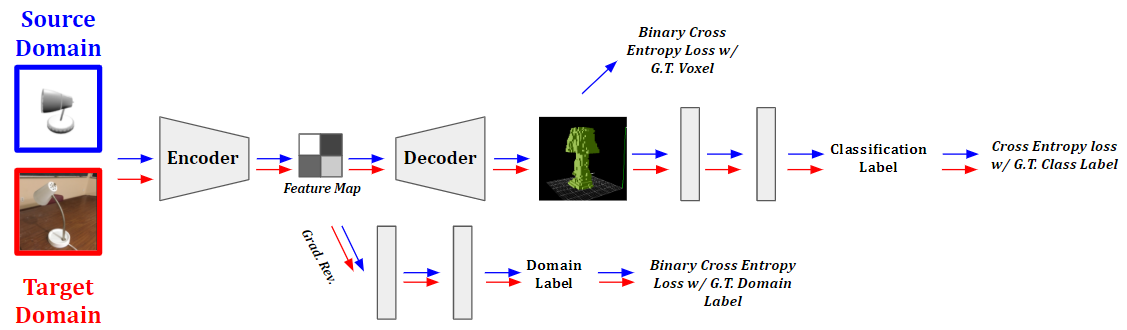} \\
\end{center}
   \caption{The proposed architecture, which is based off the pix2vox framework. We provide two additions: 1) incorporating domain adaptation in the style of DANN through a domain classifier and a gradient reversal layer, and 2) since we have classification labels for both the source and target domain, we also classify the produced voxel.}
\label{fig:model_dia}
\end{figure*}
%-------------------------------------------------------------------------
\section{Related Work} \label{sec:related}
\subsection{3D Single View Object Reconstruction}
Single view 3D reconstruction has several immediate applications. For example, it would enable objects to be reconstructed as a 3D model and placed into an augmented (AR) or virtual (VR) environment, so that the user could manipulate them in that environment. Another use case is for robotic grasping -- if an object such as a cup could be scanned, it would provide valuable information for a robot trying to pick it up by its handle. As a result, several methods have been proposed in the computer vision literature ~\cite{xie2019pix2vox, lin2018learning, choy20163d, richter2018matryoshka, tatarchenko2017octree, fan2017point}. They vary in the type of 3D representation used and each have their own trade-offs. For example, voxels are easily adapted to Convolutional Neural Networks (CNNs) but are spatially inefficient; triangular meshes are efficient but suffer from irregularities; point clouds are simple but lack explicit structural information. However, most of the these methods use ShapeNet for training and testing, and do not incorporate techniques used in the domain adaptation literature to allow for real-world viability. \cite{Pixel2Mesh} uses Graph Convolutional Neural Net (GCN) to deform a mesh of ellipsoid to obtain an output mesh. The results, however, are not very accurate and it fails to keep the genus of the ground truth. \cite{lin2018learning} uses 2D convolutional network to generate dense point clouds that shapes the surface of 3D objects in an undiscretized 3D space. The method predicts accurate shapes with higher point density but is problematic when objects contain very thin structures. \cite{choy20163d} proposed a novel architecture that unifies single and multi-view 3D reconstruction into a single framework. The method uses deep convolutional neural networks (3D Recurrent Reconstruction Neural Network) to learn a mapping from observations to their underlying 3D shapes of objects from a large collection of training data. It incrementally improves its reconstructions as it sees more views of an object but is unable to reconstruct many details and struggles with objects having high texture levels.   

\subsection{Unsupervised Domain Adaptation}
Classical approaches to unsupervised domain adaptation usually consist of matching the feature distribution between the source and target domain. Generally these methods can be categorized as either sample re-weighting (eg. \cite{4194}, \cite{jiang-zhai-2007-instance}) or feature space transformations (eg. \cite{baktashmotlagh2013unsupervised}, \cite{sun2015return}). Convolutional neural networks are also used for this purpose, because of their ability to learn powerful features. These methods, in general, are trained to minimize a classification loss while maximizing domain confusion. The classification loss is usually computed using a fully-connected or convolutional neural network trained on the labeled data. The domain confusion is usually achieved either by using a discrepancy loss, which reduces the shift between the two domains such as in (\cite{long2015learning}, \cite{pmlr-v70-long17a}, \cite{DBLP:journals/corr/BousmalisTSKE16}) or via an adverserial loss which encourages a common feature space with respect to a discriminator loss, such as in \cite{ganin2015domainadversarial}, \cite{DBLP:journals/corr/BousmalisTSKE16}, \cite{ajakan2014domainadversarial}. \cite{sun2016deep} achieves the domain-confusion by aligning the second-order statistics of the learned feature representations. In \cite{tzeng2014deep} a domain-confusion loss based on MMD \cite{gretton2008kernel} is applied to the final layer representation of a network. \cite{long2015learning} uses a sum of multiple MMDs between several layers and \cite{pmlr-v70-long17a} continues this line of work by using the joint distribution discrepancy over deep features, instead of their sum.
%-------------------------------------------------------------------------
\section{Methods} \label{sec:methods}

\subsection{3D Reconstruction Backbone Architecture}

The backbone for single view 3D reconstruction that was chosen for this project is the pix2vox architecture ~\cite{xie2019pix2vox}. It is a convolutional neural network which encodes input images into a latent feature map, which is then decoded into a $32 \times 32 \times 32$ dimensional voxel. The encoder uses convolutional layers while the decoder uses 3D transpose convolutional layers. An additional refiner CNN which is based on the U-Net is also employed to increase performance ~\cite{ronneberger2015u}. Standard techniques are used throughout the network, including batch normalization ~\cite{ioffe2015batch}, ReLU and ELU layers, and ImageNet ~\cite{deng2009imagenet} weights pretrained on VGG ~\cite{simonyan2014very}. This architecture was chosen because it is much more efficient than other competing methods such as PSGN ~\cite{fan2017point}, OGN ~\cite{tatarchenko2017octree}, and 3D-R2N2 ~\cite{choy20163d}, while being comparable or better in terms of performance. Due to limited computational resources, this was a critical factor in our decision. Note that in the original paper, pix2vox utilizes the ShapeNet dataset. 

\begin{figure*}
\begin{center}
\includegraphics[width=13cm]{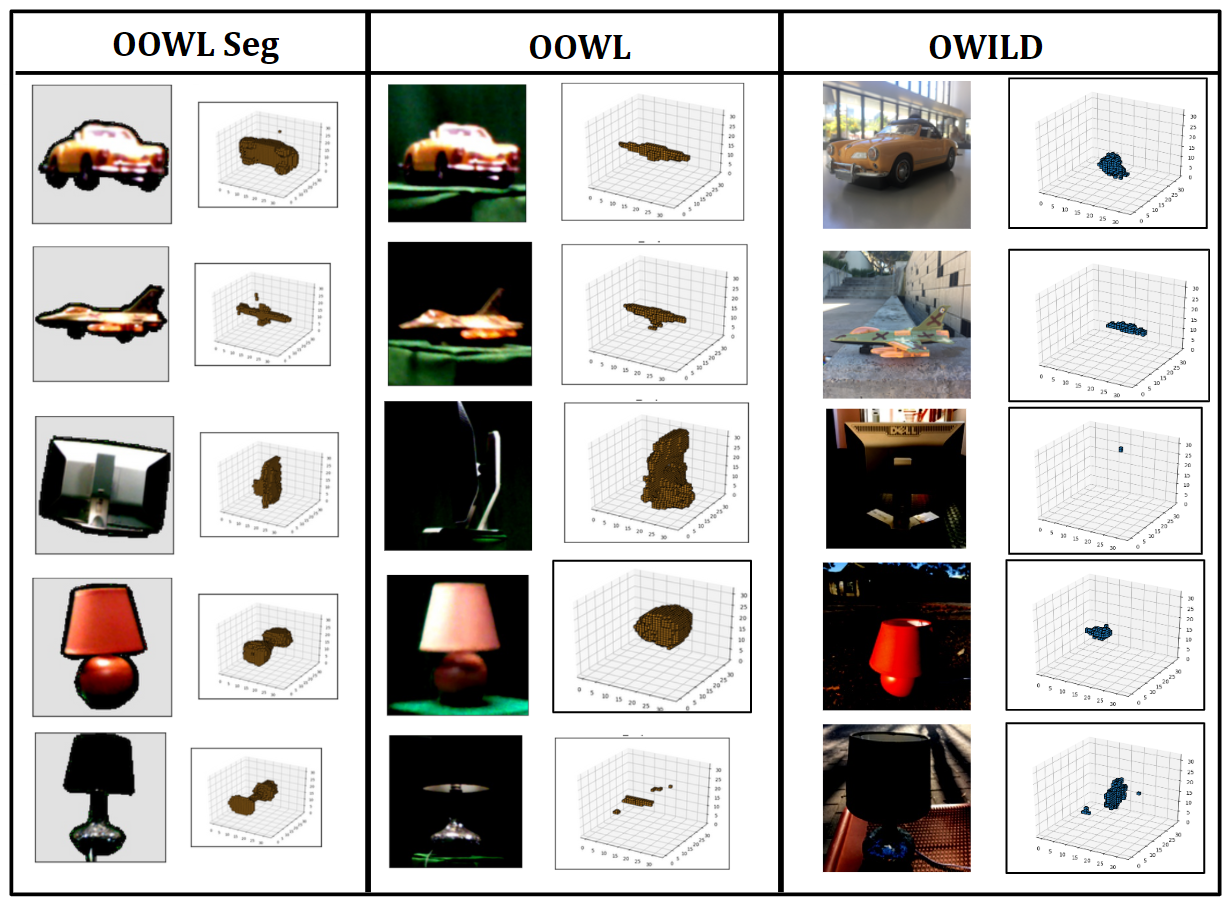} \\
\end{center}
   \caption{Reconstruction results of applying our proposed architecture on the three domains in the ODDS dataset; ShapeNet is used as the source domain, and each are trained on their respective target domains.}
\label{fig:rec_voxcls}
\end{figure*}

\subsection{Maximum Mean Discrepancy (MMD)}
Unsupervised domain adaptation is quite challenging since we do not have labeled information for the target domain. Some approaches to the problem are to try to bound the target error by the source error plus a discrepancy metric between the source and the target. The Maximum Mean Discrepancy (MMD) is a measure of the difference between two probability distributions from their samples. It is an effective criterion that compares distributions without initially estimating their density functions. Given two probability distributions $p$ and $q$ on $\mathcal{X}$, MMD is defined as
\begin{equation}
    \mathcal{MMD}(\mathcal{F},p,q) = sup_{f \in \mathcal{F}} (E_{x \sim p}[f(x)] - E_{y \sim q}[f(y)])
\end{equation}
where $\mathcal{F}$ is a class of functions $f: \mathcal{X} - \mathcal{R}$. By defining $\mathcal{F}$ as the set of functions of the unit ball in a universal Reproducing Kernel Hilbert Space (RKHS), denoted by $\mathcal{H}$, it was shown that $ \mathcal{MMD}(\mathcal{F},p,q) = 0$ will detect any discrepancy between $p$ and $q$ \cite{Kernel-MMD}.   

Let ${x_s^{(i)}}_{i=1,...n_s}$ and ${x_t^{(j)}}_{i=1,...n_t}$ be data vectors drawn from distributions $\mathcal{D}_s$ and $\mathcal{D}_t$ on the data space $\mathcal{X}$, respectively. Since $f$ is in the unit ball in a universal RKHS, we can rewrite the empirical estimate of MMD as 
\begin{equation}
    \mathcal{MMD}_{e}(x_s, x_t) = \left\Vert \frac{1}{n_s} \sum_{i=1}^{n_s} \phi(x_s^{(i)}) - \frac{1}{n_t} \sum_{j=1}^{n_t}\phi(x_t^{(j)}))  \right\Vert_{\mathcal{H}}
\end{equation}
where $\phi(\cdot): \mathcal{X} \rightarrow \mathcal{H}$ is referred to as the feature space map. 

\subsection{Deep CORAL}

Another approach to domain adaptation is by aligning the statistics of the source and target domains. CORAL \cite{sun2015return} does this by using a linear transformation to align the covariances (second order statistic) of the domains. Assuming we have a labeled source domain $\mathcal{D}_s = \{x_s^{(i)},y_s^{(i)}\}_{i=1}^{n_s}$ and an unlabeled target domain $\mathcal{D}_t = \{x_t^{(j)}\}_{j=1}^{n_t}$ where each sample is a $d$ dimensional vector, the CORAL loss is defined as:
\begin{equation}
    \ell_{CORAL} = \frac{1}{4d^2} \|C_S - C_T\|_F^2
\end{equation}
Where $\|\cdot\|_F$ is the Frobenius norm and $C_S, C_T \in \mathbb{R}^{d \times d}$ are the feature covariance matrices for the the source and target data, respectively. These matrices are given by:
\begin{equation}
    C_S = \frac{1}{n_s - 1}(D_S^TD_S - \frac{1}{n_s}(\mathbf{1}^TD_S)^T(\mathbf{1}^TD_S))
\end{equation}
\begin{equation}
    C_T = \frac{1}{n_t - 1}(D_T^TD_T - \frac{1}{n_t}(\mathbf{1}^TD_T)^T(\mathbf{1}^TD_T))
\end{equation}
Where $\mathbf{1}$ is a $d$ dimensional vector containing all $1$ and the matrices $D_S \in \mathbb{R}^{n_s \times d}, D_T\in \mathbb{R}^{n_t \times d}$ are the data matrices containing the source and target data, respectively.
%contain the $j$-th dimension of the $i$-th data sample in their $(i,j)$ position, for the source and target domains, respectively.

\begin{figure}
\begin{center}
\includegraphics[width=8cm]{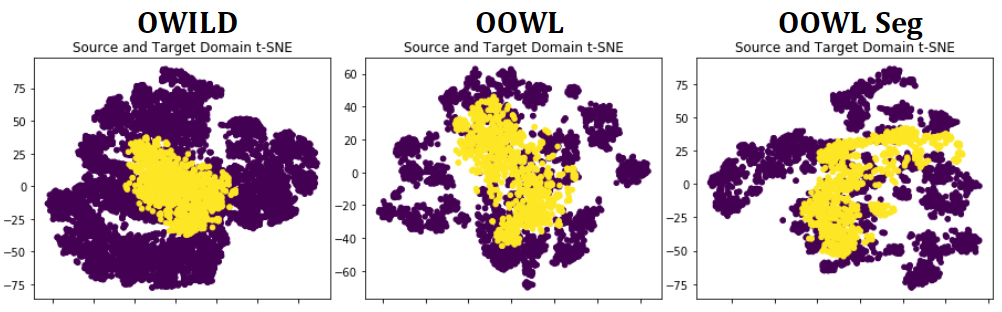} \\
\end{center}
   \caption{Learned feature map embeddings visualized with t-SNE, using our proposed model on the three target domains in the ODDS dataset. Purple denotes the source domain (ShapeNet), yellow denotes the target domain (OWILD, OOWL, and OOWL Seg). }
\label{fig:tsne_odds}
\end{figure}

\begin{figure}
\begin{center}
\includegraphics[width=8cm]{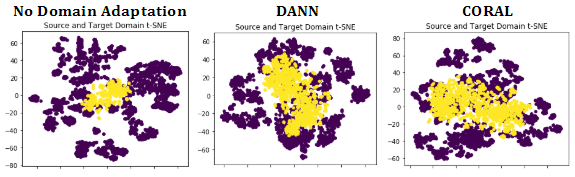} \\
\end{center}
   \caption{Learned feature map embeddings visualized with t-SNE. Purple denotes the source domain (ShapeNet), yellow denotes the target domain (OOWL). On the left we show the result when domain adaptation is not used. On the center and right, we show results achieved by applying domain adaptation (DANN or CORAL) to the vanilla pix2vox model.}
\label{fig:tsne_da}
\end{figure}

\subsection{Domain Adversarial Neural Network (DANN)}
DANN \cite{DANN} focuses on combining domain adaptation and deep feature learning under one training process. It embeds the domain adaptation method into the process of learning representation to obtain features which are discriminative and domain invariant. This is achieved by jointly optimizing the underlying features as well as two discriminative classifiers operating on these features, the \textit{label predictor and domain classifier}. The \textit{label predictor} predicts class labels and is used both during training and at test time. \textit{domain classifier} discriminates between the source and the target domains during training. The model works to \textit{minimize} the loss of the label classifier and \textit{maximize} the domain classifier loss adversarially, thereby encouraging domain-invariant features.

\subsection{Voxel Classification Architecture}
The domain adaptation techniques discussed above can be readily applied to the pix2vox architecture. However, their success may be limited, since the gap between synthetic and real domains is large. To help with this process, in addition to using domain adaptation techniques, we perform classification of the output voxel by vectorizing it, and using several fully connected layers of size 100 and 20, with ReLU activations. This is possible since we have the ground truth class labels for both the source and target domain. This proposed architecture can be seen in Figure ~\ref{fig:model_dia}; all losses are trained end-to-end. We utilize the standard cross entropy loss. This idea is inspired by the fact that in general, the output voxels should resemble their respective classes. We found that this additional source of supervision, which applies to both the source and target domain, is highly beneficial; further details can be found in Section ~\ref{sec:experiments}.

%-------------------------------------------------------------------------
\section{Experiments} \label{sec:experiments}

\subsection{Relevant Datasets}

As mentioned previously, ShapeNet is a synthetic dataset which provides many 2D rendered images of objects, as well as their ground truth 3D mesh representations (which can be converted into voxels). As shown in Figure ~\ref{fig:datasets}, they are CAD models with little detail, no texture, and no background. While the original ShapeNet has 270 classes, pix2vox uses a subset of 13 classes.

In addition, we utilize the ODDS dataset, which is a real, multiview, class-organized image dataset with multiple domains. It contains 25 classes, 6 of which overlap with the ShapeNet classes used to train the original pix2vox model. The overlapping classes are airplanes, cars, monitors, lamps, telephones, and boats, so these are the classes we use when comparing results between datasets. Each ODDS class contains 20 object instances (for example, the monitor class has 20 different types of monitors); each object instance has 8 images of it taken at 45 degree increments. Please see Figure ~\ref{fig:datasets} for some example images. There are 3 domains in the ODDS dataset that we work with. First, OWILD contains images of objects in various real-world locations, and pictures are taken with a smartphone. Second, OOWL is taken with a drone inside a lab setting. As a result, it contains several domain peculiarities such as as camera blur and a lower camera resolution. Finally, OOWLSeg is a segmented version of OOWL. Note that this data captures the real-world input statistics that we would like to work with: they're real objects, taken with smartphone cameras in various real-world locations. However, they also represent a trade-off between ease of collection and realism -- this is shown through the arrow in Figure ~\ref{fig:datasets} on the right.

\subsection{Evaluation Metrics}
The standard evaluation for 3D reconstruction when using voxels is the intersection over union (IoU) score. Formally, it is:
\begin{equation}
IoU = \frac{\sum_{i,j,k}I(p_{(i,j,k)}>t)I(gt_{(i,j,k)})}{\sum_{i,j,k}I[I(p_{(i,j,k)}>t)+I(gt_{(i,j,k)})]}
\end{equation}
where $p_{(i,j,k)}  \in [0,1]$ is the predicted occupancy probability at voxel location $(i,j,k)$ and $gt_{(i,j,k)} \in \{0,1\}$ is the ground truth value at voxel location $(i,j,k)$. Given a predicted reconstruction voxel and ground truth voxel, if $IoU=1$, then they are the same voxel. If $IoU = 0$, then there is no intersection between the two voxels. Thus, a higher IoU score indicates a better reconstruction result. It is important to note that for the case of the ODDS dataset, we do not have ground truth (currently, we are not aware of any publicly available, sufficiently large multiview dataset with 3D ground truth). Therefore, for the scope of this paper, it is primarily used as a test dataset to judge qualitative reconstruction results. We cannot quantitatively evaluate metrics like IoU, due to the lack of 3D ground truth (it is unsupervised in this regard). However, we do try to utilize the ground truth class labels that come with OWILD as supervision.

\subsection{Application of Domain Adaptation on a Vanilla Pix2Vox Model}
\begin{figure}
\begin{center}
\includegraphics[width=8cm]{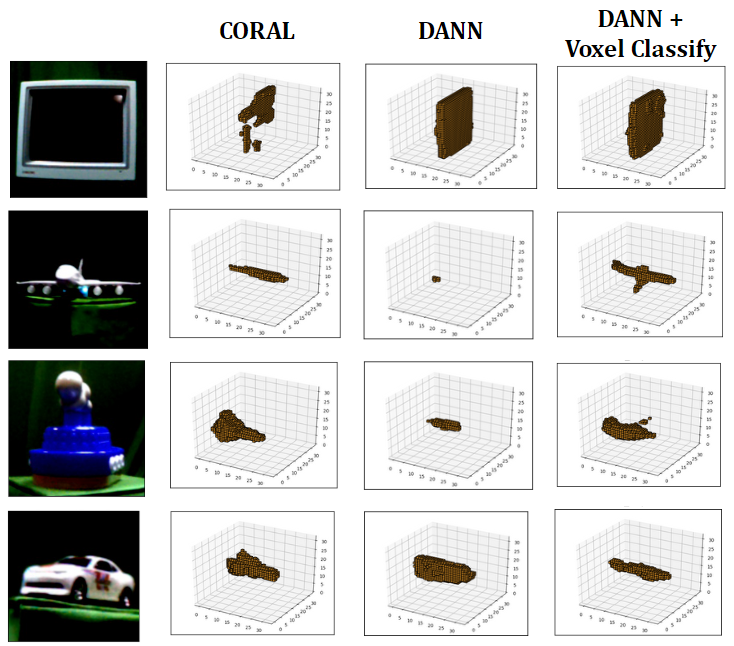} \\
\end{center}
   \caption{Reconstruction results when domain adaptation is used with OOWL as the target domain. We compare domain adaptation using Deep CORAL, DANN, and our proposed architecture (DANN + Voxel Classify).}
\label{fig:rec_da}
\end{figure}

First, we report the results of applying DANN and Deep CORAL domain adaptation to the vanilla pix2vox model for single view 3D reconstruction. Qualitative reconstruction results are shown in Figure \ref{fig:rec_da}, under ``CORAL'' and ``DANN''. We observed that reconstruction when not using any domain adaptation generally looks like random noise (these reconstruction results are omitted to save space). Also, in our experiments we found that MMD was not effective; therefore, MMD results have also been omitted. Regarding Deep CORAL and DANN, we can see that in general both help, though results are still far from perfect. Visually checking the reconstruction results, we found that DANN performed better than CORAL.  We also embed the learned feature maps into 2D, using t-SNE as the dimensionality reduction algorithm. This is shown in Figure \ref{fig:tsne_da}. We can see that the use of DANN and Deep CORAL both helps to make embedded features more domain invariant -- the distributions of the source ShapeNet domain (purple) are more matched with the distributions of the target OOWL domain (yellow). We also note that the introduction of domain adaptation also negatively impacts the IoU on the source domain -- this is shown in Table \ref{tab:iou_results}. Intuitively, this makes sense since the network is constrained to only output domain invariant latent representations. This makes training more difficult. In the future, we would like to look into this more and see if we can maintain IoU results on the source domain while performing domain adaptation.

\subsection{Reconstruction with a Voxel Classification Loss}
\begin{figure}
\begin{center}
\includegraphics[width=8cm]{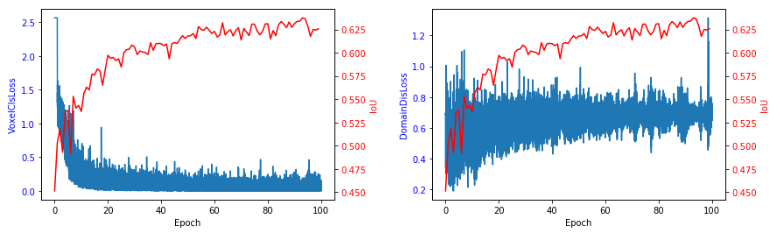} \\
\end{center}
   \caption{Training results on our architecture. On the left, the voxel classification loss (in blue) is plotted against the IoU. On the right, the domain discrepancy loss (in blue) is plotted against the IoU.}
\label{fig:training}
\end{figure}

As mentioned above, we found that regarding the vanilla pix2vox model, DANN is helpful. However, results are still sub optimal. To address this, we utilize our proposed voxel classification network. Training is performed end-to-end, and we report training losses as a function of epoch in Figure \ref{fig:training}. While the voxel classification loss does decrease, during training we found it difficult to reduce it past epoch 20 -- it fluctuates beyond that point. In the future, we plan on trying to look into ways of address this. Meanwhile, the domain discrepancy loss is maintained at around 0.5. This is expected due to adversarial training induced by the gradient reversal layer.

Next, using this trained model we evaluate the differences between the domains in ODDS. This gives us a way to see how large the domain gap is between Shapenet and the target domains OOWL, OOWL Seg, and OWILD. We report t-SNE embeddings in Figure \ref{fig:tsne_odds} and the reconstructions in Figure \ref{fig:rec_voxcls} for the three target domains. We can see that in general, it appears that OWILD is the most challenging dataset. We believe that this is because OWILD has complex backgrounds, which make it very far from the shapenet domain. On the other hand, we can see that OOWL seg performs quite well. We believe that the segmentation makes the images very similar to ShapeNet, which also do not have a background. 

\begin{table}
\begin{center}
\begin{tabular}{|c|c|c|c|}
\hline
Class & No DA & DANN & CORAL \\
\hline\hline
Airplane & 0.6842 & 0.6046 & 0.6377 \\
Car & 0.8548 & 0.8377 & 0.8485 \\
Monitor & 0.5373 & 0.4845 & 0.4968 \\
Lamp & 0.4430 & 0.4228 & 0.4322 \\
Telephone & 0.7764 & 0.7278 &  0.7555\\
Boat & 0.5946 & 0.5852 &  0.5926 \\
\hline
Overall & 0.7110 & 0.6762 & 0.6925 \\
\hline
\end{tabular}
\end{center}
\caption{IoU results for a model trained on ShapeNet with and without OOWL domain adaptation (DANN, Deep CORAL). Results reported for $t=0.4$.}
\label{tab:iou_results}
\end{table}

%-------------------------------------------------------------------------
\section{Conclusion and Future Work}

In this paper, we have focused on the task of single view voxel reconstruction in the real world. To do this, we extended the pix2vox architecture using domain adaptation between the supervised synthetic ShapeNet dataset and the unsupervised, real ODDS dataset. However, we showed that simply applying domain adaptation is not enough; reconstruction results are only marginally better. Therefore, we proposed an architecture which also utilizes a voxel classification loss in addition to an adversarial loss, which led to better results.

There are several extensions that were not done due to time and computational constraints which we plan on exploring in the future. First, it would be interesting to see if our conclusions in the project hold for other architectures (eg mesh or point clouds). Second, no large dataset exists with real world ground truth 3D data. Perhaps the closest is Pix3D, but it only has 8 classes, and pictures are only from one angle. No class overlaps with Pix3D, OWILD, and ShapeNet. If such a dataset existed, it would be feasible to achieve quantitative, not just qualitative reconstruction results. Third, we plan on working towards improving results on OWILD through more experimentation. For example, we want to try data augmentation on ShapeNet by pasting background from the MIT Places datasets \cite{zhou2017places} and see if that improves results. We also want to further explore the domain gaps between the datasets through methods like domain bridges, which proposes domain adaptation gradually over several intermediate domains which increase in difficulty to the final target domain \cite{dai2019adaptation}. Finally, because ODDS is a multiview dataset, it would be natural to generalize results to the multiview reconstruction case. \\

{\small
\bibliographystyle{ieee}
\bibliography{egbib}
}

\end{document}